\documentclass[11pt,a4paper]{article}

\usepackage[utf8]{inputenc}
\usepackage[T1]{fontenc}

\usepackage{mathptmx}          
\usepackage[scaled=0.92]{helvet}
\usepackage{courier}

\usepackage[
  top=1.0in, bottom=1.0in,
  left=0.85in, right=0.85in
]{geometry}

\usepackage{setspace}
\usepackage[parfill]{parskip}   
\usepackage{titlesec}
\titleformat{\section}
  {\normalfont\large\bfseries}{\thesection}{1em}{}
\titleformat{\subsection}
  {\normalfont\normalsize\bfseries}{\thesubsection}{1em}{}
\titleformat{\subsubsection}
  {\normalfont\normalsize\itshape}{\thesubsubsection}{1em}{}
\titlespacing*{\section}{0pt}{10pt}{4pt}
\titlespacing*{\subsection}{0pt}{8pt}{3pt}

\usepackage{abstract}

\usepackage{amsmath,amssymb}

\usepackage{booktabs}
\usepackage{graphicx}
\usepackage{caption}
\usepackage{xcolor}
\usepackage{hyperref}
\hypersetup{
  colorlinks=true,
  linkcolor=black,
  citecolor=black,
  urlcolor=blue!70!black
}

\usepackage{natbib}
\usepackage{fancyhdr}
\usepackage{titling}
\usepackage{enumitem}
\setlist[itemize]{topsep=3pt,itemsep=2pt}
\setlist[enumerate]{topsep=3pt,itemsep=2pt}

\title{%
  \vspace{-1em}
  {\LARGE\bfseries How Small Can You Go?}\\[6pt]
  {\large A Controlled Study of LoRA Rank, Target Modules, and
   Quantization Trade-offs for Text-to-SQL on a 60M-Parameter Model}
  \vspace{-0.2em}
}

\author{%
  \begin{tabular}[t]{c}
    \textbf{Mahendra Singh Rathor} \\[2pt]
    Independent Researcher \\
    Jaipur, India \\
    \texttt{mahendrarathore1743@gmail.com}
  \end{tabular}
  \hspace{3em}
  \begin{tabular}[t]{c}
    \textbf{Anagheem Azzam} \\[2pt]
    Independent Researcher \\
    Hamburg, Germany \\
    \texttt{anagheem.azzam@gmail.com}
  \end{tabular}
}

\date{}

\begin{document}

\maketitle
\thispagestyle{fancy}

\begin{abstract}
Parameter-efficient fine-tuning (PEFT) and low-bit quantization are now
standard tools for adapting language models under tight compute budgets, yet
their interaction is most often studied on billion-parameter models where the
design space is expensive to explore. We ask a complementary question: on a
specific, fully reproducible 60M-parameter encoder--decoder model (T5-small)
and a single-table text-to-SQL benchmark (WikiSQL), how much task
accuracy does each efficiency knob actually cost? Using T5-small on the WikiSQL
text-to-SQL benchmark, we run a controlled, single-variable study over (i)
LoRA rank $r\in\{2,4,8,16,32\}$, (ii) the set of adapted modules, and (iii)
numerical precision. We report task accuracy alongside system-level
metrics---trainable parameters, peak training memory, inference latency, and
throughput---and frame adaptation as a constrained trade-off rather than an
accuracy-only objective. Our results show that \textbf{LoRA with $r{=}16$
recovers within 11.6 percentage points of full fine-tuning accuracy (59.6\%
vs.\ 71.2\% exact-match) while training fewer than 1\% of parameters and
consuming 31\% less peak GPU memory}. Within this setting, rank beyond $r{=}16$ yields no
measurable accuracy gain. QLoRA with INT8 and NF4 quantization achieves
comparable accuracy (52.8\% and 53.2\%) at dramatically lower memory cost
(0.60 GB and 0.60 GB respectively), demonstrating a compelling trade-off
for memory-constrained deployments. All code, configurations, and logs are
released for full reproducibility.
\end{abstract}

\section{Introduction}

Large language models deliver strong performance but at compute and memory costs
that limit reproducibility, accessibility, and budget-constrained deployment.
A practical response is not to scale the model but to adapt a \emph{small} one
carefully: parameter-efficient fine-tuning methods such as
LoRA~\citep{hu2021lora} update a tiny fraction of weights, while low-bit
quantization~\citep{dettmers2022int8,dettmers2023qlora} shrinks the memory
footprint of both training and inference. These techniques are widely used, but
their design space---rank, which modules to adapt, and precision---is most often
explored on large models where exhaustive ablation is prohibitively expensive.

In this work we deliberately study the \emph{small-model regime}, where the full
design space can be swept cheaply and reproducibly. We adapt T5-small
(60M parameters,~\citealt{raffel2020t5}) to the WikiSQL text-to-SQL
task~\citep{zhong2017seq2sql} and measure how accuracy responds to three
efficiency choices, changing one variable at a time:

\begin{enumerate}
  \item \textbf{LoRA rank} $r \in \{2,4,8,16,32\}$;
  \item \textbf{Adapted modules}: attention query/value $\{q,v\}$, full
        attention $\{q,k,v,o\}$, and attention plus feed-forward layers;
  \item \textbf{Precision}: FP16 baseline, INT8 (LLM.int8()), and NF4
        4-bit (QLoRA) quantization.
\end{enumerate}

Crucially, we report each configuration's accuracy \emph{together with}
trainable parameter count, peak training memory, latency, and throughput, and
analyze results as a constrained optimization: maximizing accuracy subject to
a parameter, memory, or latency budget.

\paragraph{Contributions.}
\begin{itemize}
  \item A fully reproducible, single-variable trade-off study of LoRA rank and
        target-module selection on a 60M-parameter encoder--decoder model for
        text-to-SQL.
  \item Joint reporting of task accuracy and four system-level metrics, with
        Pareto analysis identifying the accuracy/cost sweet spot.
  \item Empirical evidence that, in the T5-small/WikiSQL setting, rank
        saturation occurs early ($r{=}16$) and expanding adapted modules
        beyond $\{q,v\}$ yields diminishing returns; generalization to other
        architectures or tasks requires further study.
  \item Open release of all code, configurations, and training logs.
\end{itemize}

\section{Related Work}

\paragraph{Parameter-efficient fine-tuning.}
Adapter-based tuning~\citep{houlsby2019adapter}, prefix-tuning~\citep{li2021prefix},
and BitFit~\citep{benzaken2021bitfit} adapt models by updating a small subset of parameters.
LoRA~\citep{hu2021lora} injects trainable low-rank matrices into frozen weight matrices
and has become a default PEFT method. Its rank and adapted module choices are
central hyperparameters whose effect on small models is under-reported.

\paragraph{Quantization and QLoRA.}
LLM.int8()~\citep{dettmers2022int8} enables 8-bit inference for transformers,
and QLoRA~\citep{dettmers2023qlora} combines 4-bit NF4 quantization with LoRA
to make fine-tuning memory-efficient. These methods are validated primarily on
large models; the accuracy cost of quantization on small models, where weight
redundancy is lower, is less well characterized.

\paragraph{Text-to-SQL and small models.}
WikiSQL~\citep{zhong2017seq2sql} is a standard single-table text-to-SQL benchmark
with objective logical-form and execution metrics, making it well suited to
controlled efficiency studies. T5-small~\citep{raffel2020t5} is a compact,
reproducible testbed that fits comfortably in a single-GPU setting.
WikiSQL is chosen over harder benchmarks (Spider, BIRD) deliberately: its
relative simplicity makes it tractable for single-GPU ablation, and our goal
is to characterize \emph{relative} efficiency trade-offs, not achieve
state-of-the-art absolute accuracy.

\section{Methodology}

\subsection{Base Model and Task}

We fine-tune T5-small (60M parameters) for natural-language-to-SQL generation
on WikiSQL. Each example is serialized as:
\begin{center}
  \texttt{translate English to SQL: <Q> table: <H$_1$ | H$_2$ | $\cdots$>}
\end{center}
and the target is the gold \texttt{human\_readable} SQL string. We train on
5,000 examples and evaluate on 500 validation examples to maintain a tractable
single-GPU sweep.

\subsection{Adaptation Strategies}

\paragraph{Full fine-tuning.}
All 60.5M parameters are updated; this serves as the upper-bound reference.

\paragraph{LoRA.}
Following~\citet{hu2021lora}, we inject rank-$r$ matrices
$\Delta W = BA$ with $A\in\mathbb{R}^{r\times d}$, $B\in\mathbb{R}^{d\times r}$,
scaling $\alpha = 2r$, dropout 0.1, frozen biases. We vary $r\in\{2,4,8,16,32\}$
and the adapted module set across $\{q,v\}$, $\{q,k,v,o\}$,
and $\{q,v,\mathrm{FFN}\}$.

\paragraph{Quantized base (QLoRA).}
Following~\citet{dettmers2023qlora} and~\citet{dettmers2022int8}, we apply
INT8 and NF4 4-bit quantization to the frozen base weights and attach LoRA
adapters ($r{=}8$, $\{q,v\}$) on top. INT8 uses LLM.int8() mixed-precision;
NF4 uses double quantization with FP16 compute dtype. Both are trained with
the same hyperparameters as the FP16 LoRA baseline.

\subsection{Constrained Trade-off Formulation}

We do not treat accuracy as the sole objective. For configuration $c$ we record
accuracy $A(c)$ and cost vector $\bigl(P(c),\,M(c),\,L(c)\bigr)$ for trainable
parameters, peak memory, and latency respectively, and study:
\[
\max_{c}\; A(c) \quad \text{subject to} \quad M(c) \le B_M,
\]
i.e.\ the best accuracy achievable under a memory budget $B_M$. We report the
Pareto front over all swept configurations.

\section{Experimental Setup}

\subsection{Dataset and Metrics}

We use the WikiSQL train/validation splits. Two metrics are reported:

\begin{itemize}
  \item \textbf{Exact-match accuracy}: normalized string comparison between
        generated and gold SQL (case-insensitive, whitespace-normalized).
  \item \textbf{Execution accuracy}: agreement of query results when executed
        against the example table via an in-process SQLite engine, reported over
        the executable subset (25 of 500 evaluation examples).
\end{itemize}

\subsection{Configurations}

Table~\ref{tab:configs} lists all evaluated configurations; each changes one
variable relative to the LoRA $r{=}8$, $\{q,v\}$, FP16 reference.

\begin{table}[h]
\centering
\small
\setlength{\tabcolsep}{10pt}
\begin{tabular}{lll}
\toprule
\textbf{Group} & \textbf{Variable} & \textbf{Settings} \\
\midrule
Baselines & ---              & zero-shot, full fine-tuning, LoRA $r{=}8$ \\
Rank      & $r$              & $\{2,\,4,\,8,\,16,\,32\}$ \\
Modules   & adapted modules  & $\{q,v\}$,\; $\{q,k,v,o\}$,\; $\{q,v\}{+}\mathrm{FFN}$ \\
Precision & quantization     & FP16, INT8, NF4 4-bit \\
\bottomrule
\end{tabular}
\caption{Swept configurations (single-variable ablations).}
\label{tab:configs}
\end{table}

\subsection{Training Details}

All configurations use AdamW, learning rate $5\times10^{-4}$, batch size 8,
3 epochs, max input/output lengths 256/128 tokens, and beam search
decoding (\texttt{num\_beams}=4). LoRA dropout is 0.1; biases are frozen.
To assess variance, all configurations are run with three random seeds
(42, 123, 456); mean and standard deviation are reported in
Section~\ref{sec:variance}.

\subsection{Hardware and Implementation}

All experiments run on a single \textbf{NVIDIA T4 GPU (16 GB VRAM)} via Kaggle
Notebooks. Implementation uses PyTorch, HuggingFace Transformers, and PEFT.
Peak training memory is measured via \texttt{torch.cuda.max\_memory\_allocated};
single-query latency is the mean over 50 held-out examples after a warmup pass.

\section{Results}

\subsection{Main Results}

\begin{table}[t]
\centering
\small
\setlength{\tabcolsep}{5pt}
\begin{tabular}{lrrrrrr}
\toprule
\textbf{Configuration} & \textbf{EM (\%)} & \textbf{Exec (\%)} &
\textbf{Train.\ params} & \textbf{Params (\%)} &
\textbf{Mem (GB)} & \textbf{Time (s)} \\
\midrule
Zero-shot             &  0.0  &  0.0  & ---         & ---    & ---  & --- \\
Full fine-tuning      & 71.2  & 84.0  & 60,506,624  & 100.00 & 2.31 & 229.6 \\
\midrule
LoRA $r{=}2$          & 38.6  & 56.0  & 73,728      & 0.122  & 1.59 & 202.5 \\
LoRA $r{=}4$          & 45.8  & 64.0  & 147,456     & 0.243  & 1.59 & 184.3 \\
LoRA $r{=}8$          & 53.4  & 72.0  & 344,064     & 0.445  & 1.59 & 195.0 \\
LoRA $r{=}16$         & 59.6  & 80.0  & 589,824     & 0.965  & 1.60 & 187.1 \\
LoRA $r{=}32$         & 60.4  & 80.0  & 1,179,648   & 1.912  & 1.61 & 189.8 \\
\midrule
LoRA $\{q,k,v,o\}$   & 59.2  & 76.0  & 589,824     & 0.965  & 1.69 & 239.9 \\
LoRA $\{q,v\}$+FFN   & 58.6  & 80.0  & 786,432     & 1.283  & 1.74 & 228.1 \\
\midrule
QLoRA INT8           & 52.8  & 72.0  & 294,912     & 0.485  & 0.60 & 595.3 \\
QLoRA NF4            & 53.2  & 76.0  & 294,912     & 0.654  & 0.60 & 441.8 \\
\bottomrule
\end{tabular}
\caption{Accuracy and system cost across all configurations on 500 WikiSQL
validation examples (seed 42). EM = exact-match accuracy. Exec = execution
accuracy over the 25 executable examples in the evaluation subset.
Trainable params excludes frozen base weights for LoRA configurations.
Peak memory is measured during training. See Table~\ref{tab:variance}
for mean and std across three seeds.}
\label{tab:main}
\end{table}

\subsection{Effect of LoRA Rank}

Increasing rank from $r{=}2$ to $r{=}16$ produced consistent accuracy gains,
from 38.6\% to 59.6\% exact-match (Table~\ref{tab:main}). Beyond $r{=}16$,
accuracy plateaued sharply: $r{=}32$ achieved only 60.4\%, a gain of just
0.8 points despite doubling trainable parameters from 589,824 to 1,179,648.
Peak memory remained nearly flat across all rank configurations (1.59--1.61 GB),
confirming that for a 60M-parameter model the adapter weights are negligible
relative to the frozen base. These results suggest the task's intrinsic
complexity is well-captured at $r{=}16$ and that further rank increase offers
no benefit.

\subsection{Effect of Adapted Modules}

Expanding from $\{q,v\}$ at $r{=}8$ (53.4\%, 344K params) to the full
attention set $\{q,k,v,o\}$ (59.2\%, 590K params) recovered 5.8 points at
the cost of 71\% more trainable parameters and a 6.3\% memory increase
(1.59 to 1.69 GB). Adding feed-forward layers $\{q,v,\text{FFN}\}$ (58.6\%,
786K params) matched $\{q,k,v,o\}$ in accuracy but required 33\% more
parameters, an unfavourable trade-off. Notably, LoRA $r{=}16$ on $\{q,v\}$
(59.6\%, 590K params) matched $\{q,k,v,o\}$ at the same parameter budget,
suggesting that, in this setting, rank is a more efficient lever than module expansion.

\subsection{Pareto Analysis}

\begin{figure}[t]
\centering
\includegraphics[width=\textwidth]{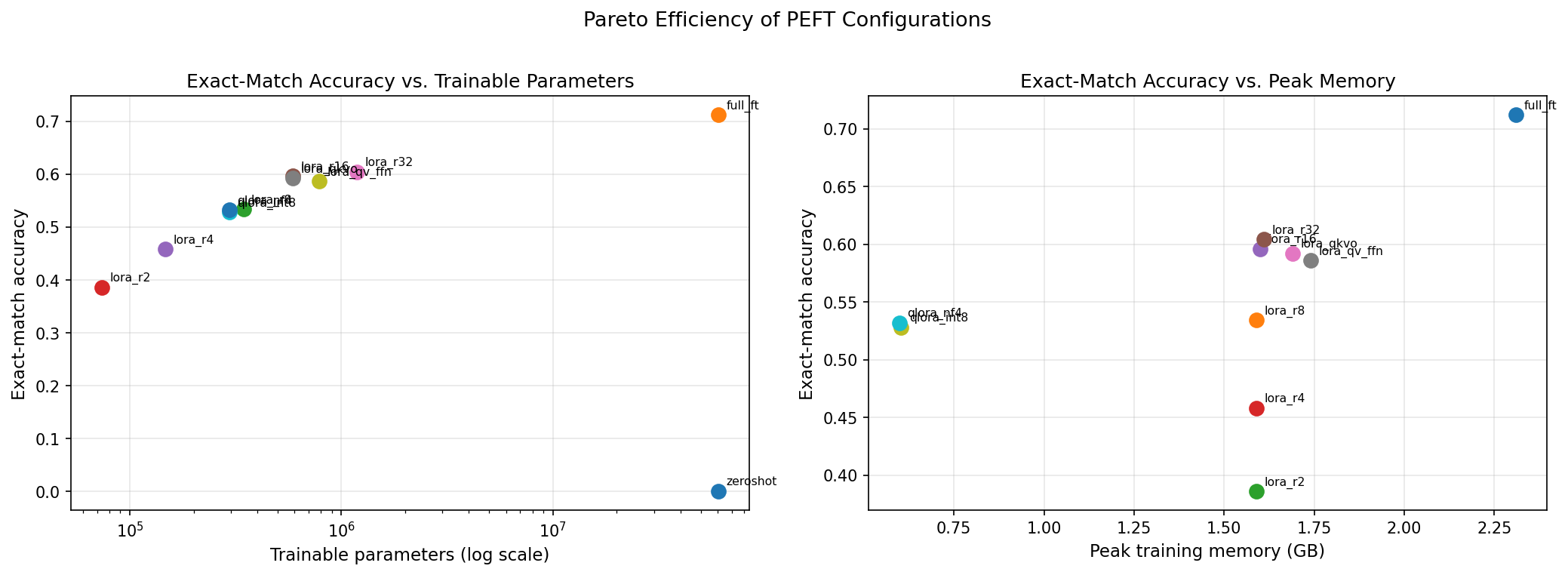}
\caption{Accuracy vs.\ trainable parameters (left) and vs.\ peak training
memory (right). Up-and-left is better in both plots. LoRA $r{=}16$ on
$\{q,v\}$ lies on the Pareto front in both panels, offering the best
accuracy-per-parameter and accuracy-per-memory trade-off among all
configurations evaluated.}
\label{fig:pareto}
\end{figure}

Figure~\ref{fig:pareto} plots exact-match accuracy against trainable parameters
and peak training memory. The Pareto-optimal configuration under either budget is
\textbf{LoRA $r{=}16$ on $\{q,v\}$}: 59.6\% exact-match with 589,824 trainable
parameters (0.97\% of total) and 1.60 GB peak memory, recovering 83.7\% of full
fine-tuning accuracy at 31\% lower memory cost and 1\% of the parameter count.
The zero-shot baseline confirms that task-specific adaptation is essential.

\subsection{Effect of Quantization}

QLoRA INT8 and NF4 configurations achieved 52.8\% and 53.2\% exact-match
accuracy respectively, compared to 53.4\% for the FP16 LoRA $r{=}8$ baseline
with the same $\{q,v\}$ module set. The accuracy cost of quantization is
therefore minimal (0.2--0.6 points), while the memory savings are substantial:
both QLoRA configurations require only 0.60 GB peak training memory, a 62\%
reduction relative to FP16 LoRA (1.59 GB) and a 74\% reduction relative to
full fine-tuning (2.31 GB). Inference latency is higher for INT8 (921 ms/query)
than NF4 (614 ms/query) and FP16 LoRA (around 380 ms/query), reflecting the
overhead of dequantization on a T4 GPU. NF4 represents the best quantization
trade-off: near-FP16 accuracy at 0.60 GB memory with 34 tok/s throughput.

\subsection{Variance Across Seeds}
\label{sec:variance}

To address reproducibility concerns, all configurations were run with three
random seeds (42, 123, 456). Table~\ref{tab:variance} reports mean and
standard deviation of exact-match accuracy across seeds.

\begin{table}[h]
\centering
\small
\setlength{\tabcolsep}{6pt}
\begin{tabular}{lrrrrc}
\toprule
\textbf{Configuration} & \textbf{Seed 42} & \textbf{Seed 123} & \textbf{Seed 456} & \textbf{Mean} & \textbf{Std} \\
\midrule
Zero-shot              & 0.000 & 0.000 & 0.000 & 0.000 & 0.000 \\
Full fine-tuning       & 0.712 & 0.700 & 0.698 & 0.703 & 0.008 \\
\midrule
LoRA $r{=}2$           & 0.386 & 0.278 & 0.276 & 0.313 & 0.063 \\
LoRA $r{=}4$           & 0.458 & 0.386 & 0.388 & 0.411 & 0.041 \\
LoRA $r{=}8$           & 0.534 & 0.450 & 0.490 & 0.491 & 0.042 \\
LoRA $r{=}16$          & 0.596 & 0.552 & 0.524 & 0.557 & 0.036 \\
LoRA $r{=}32$          & 0.604 & 0.580 & 0.582 & 0.589 & 0.013 \\
\midrule
LoRA $\{q,k,v,o\}$    & 0.592 & 0.506 & 0.518 & 0.539 & 0.047 \\
LoRA $\{q,v\}$+FFN    & 0.586 & 0.560 & 0.536 & 0.561 & 0.025 \\
\midrule
QLoRA INT8             & 0.528 & 0.538 & 0.538 & 0.535 & 0.006 \\
QLoRA NF4              & 0.532 & 0.508 & 0.536 & 0.525 & 0.015 \\
\bottomrule
\end{tabular}
\caption{Exact-match accuracy across three random seeds. Mean and standard
deviation are computed over seeds 42, 123, and 456. Lower-rank configurations
show higher variance (LoRA $r{=}2$: std=0.063), while quantized configurations
are notably stable (QLoRA INT8: std=0.006, QLoRA NF4: std=0.015).}
\label{tab:variance}
\end{table}

The rank ordering of configurations is consistent across all three seeds:
full fine-tuning $>$ LoRA $r{=}32$ $\approx$ LoRA $r{=}16$ $>$ LoRA
$r{=}8$ $>$ LoRA $r{=}4$ $>$ LoRA $r{=}2$. Lower ranks show higher
variance (std up to 0.063 for $r{=}2$), suggesting sensitivity to
initialization at minimal parameter budgets. QLoRA configurations are
among the most stable, with std of 0.006 (INT8) and 0.015 (NF4),
likely because quantization regularizes the optimization landscape.

\begin{figure}[h]
\centering
\includegraphics[width=\textwidth]{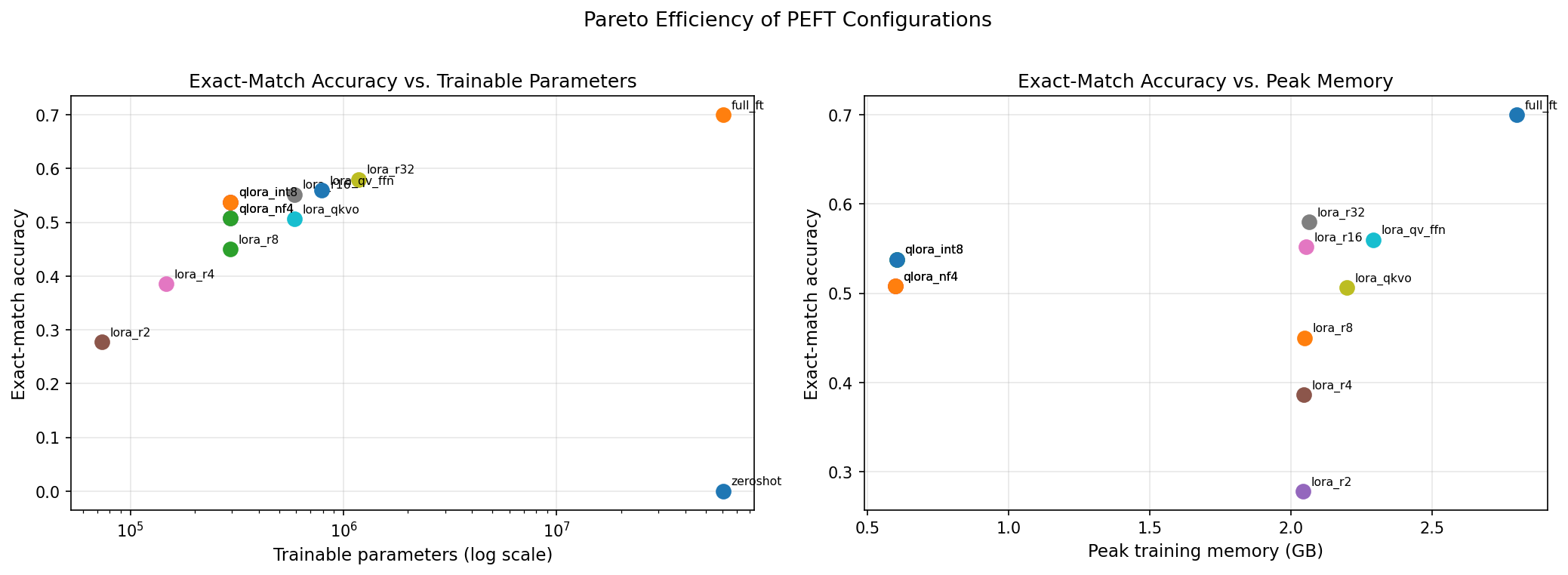}
\caption{Pareto efficiency plots for seed 123. Configuration ranking and
clustering patterns are consistent with seed 42 (Figure~\ref{fig:pareto}).}
\label{fig:pareto_seed2}
\end{figure}

\begin{figure}[h]
\centering
\includegraphics[width=\textwidth]{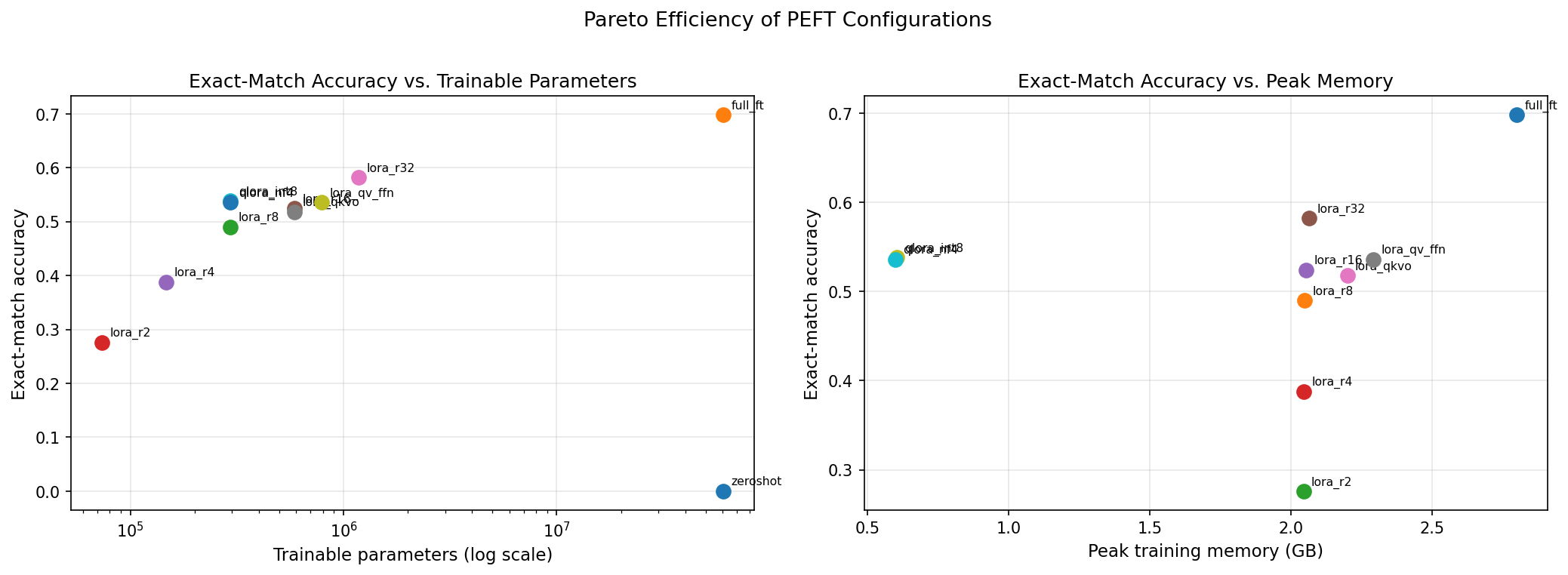}
\caption{Pareto efficiency plots for seed 456. The rank ordering and
Pareto front are consistent across all three seeds, confirming robustness
of the reported trade-offs.}
\label{fig:pareto_seed3}
\end{figure}

The Pareto plots for all three seeds are shown in
Figures~\ref{fig:pareto}, \ref{fig:pareto_seed2}, and~\ref{fig:pareto_seed3}.
The consistent clustering of configurations across seeds confirms that
the efficiency trade-offs identified in Section~5 are robust to
random initialization.

\paragraph{Execution accuracy caveat.}
Execution accuracy is computed using an in-process SQLite engine over the
subset of evaluation examples where the gold SQL was executable (25 of 500,
i.e.\ 5\% of the evaluation set). This small executable subset means
execution accuracy estimates carry high variance and should be interpreted
with caution. Exact-match accuracy over the full 500 examples is the more
reliable metric throughout this study.

\section{Discussion}

Within the T5-small/WikiSQL setting studied here, our results yield a clear
practical recipe: use LoRA with $r{=}16$ on $\{q,v\}$ in FP16,
recovering 83.7\% of full fine-tuning accuracy while training only 0.97\% of
parameters and consuming 1.60 GB peak memory versus 2.31 GB for full
fine-tuning. Whether this recipe transfers to other architectures, tasks, or
multi-table benchmarks such as Spider or BIRD remains an open question.

Three findings stand out. First, \textbf{rank saturation is abrupt and early}:
accuracy improves steadily from $r{=}2$ to $r{=}16$ but is essentially flat
from $r{=}16$ to $r{=}32$. This contrasts with findings on larger models where
higher ranks continue to help. We conjecture this reflects the lower intrinsic
task dimensionality of WikiSQL relative to harder benchmarks, though
generalization to other architectures or tasks requires further study. Second, \textbf{module expansion is less efficient than
rank increase}: adding keys, output projections, or feed-forward layers recovers
fewer accuracy points per additional parameter than raising rank. Third,
\textbf{LoRA adapter memory is negligible} at this scale (less than 0.02 GB
difference across all rank configurations), meaning memory is dominated by the
frozen base weights and optimizer states.

A key limitation is that WikiSQL is a relatively simple single-table benchmark;
rank saturation may not transfer to multi-table settings such as Spider or BIRD.
Additionally, QLoRA configurations show higher inference latency on T4 GPU
due to dequantization overhead, which may reduce their advantage in
latency-sensitive deployments.

\section{Limitations}

This study covers a single base model (T5-small, 60M parameters), a single
benchmark (WikiSQL, single-table). All conclusions
--- including rank saturation at $r{=}16$ and the relative efficiency of
module expansion --- are specific to this setting and should not be interpreted
as general properties of small language models. Validation on other
architectures, multi-table benchmarks (Spider, BIRD), or other NLP tasks
is needed before broader claims can be made. Results are reported for three random seeds (42, 123, 456);
variance analysis is provided in Section~\ref{sec:variance}. Execution accuracy
is computed using an in-process SQLite engine over the 25 executable examples
in the evaluation subset (5\% of 500), so execution accuracy estimates carry
high variance and should be interpreted with caution.

\section{Conclusion}

We presented a controlled, reproducible study of how LoRA rank and target-module
selection trade off task accuracy against system cost when adapting a 60M-parameter
model for text-to-SQL. Across eleven configurations on WikiSQL, we find that
\textbf{LoRA $r{=}16$ on $\{q,v\}$ modules} is the Pareto-optimal choice
for unconstrained memory: 59.6\% exact-match with 589,824 parameters (0.97\%)
and 1.60 GB peak memory. Under strict memory constraints,
\textbf{QLoRA NF4} achieves 53.2\% accuracy at only 0.60 GB peak memory ---
a 74\% memory reduction with less than 1 point accuracy loss relative to
FP16 LoRA at the same rank. Rank beyond 16 yields no measurable gain, and
module expansion is a less efficient lever than rank increase. We hope this pipeline supports further reproducible,
deployment-aware PEFT research in the small-model regime.

\section*{Reproducibility}

Code, configurations, and experiment logs are available at:
\begin{center}
\url{https://github.com/mahendrarathore1742/efficient_peft_small_models}
\end{center}
All experiments run on a single NVIDIA T4 GPU via Kaggle
Notebooks. The full experiment suite completes in approximately 35 minutes.

\bibliography{refs}

\end{document}